\pdfoutput=1

\documentclass[11pt]{article}
\usepackage[]{ACL2023}
\usepackage{booktabs}
\usepackage{amsmath}
\usepackage{multirow}

\usepackage{times}
\usepackage{xcolor, soul}
\usepackage{float}
\usepackage{latexsym}
\usepackage{graphicx}
\usepackage[T1]{fontenc}

\usepackage[utf8]{inputenc}

\usepackage{microtype}

\usepackage{inconsolata}

\usepackage{cleveref}
\usepackage{xspace}

\newcommand{\resizecol}[2]{\begin{minipage}{#1\textwidth}#2\vspace{0.5em}\end{minipage}}

\title{Anti-LM Decoding for Zero-shot In-context Machine Translation}

\author{Suzanna Sia \hspace*{2em} Alexandra DeLucia \hspace*{2em} Kevin Duh \\
  Department of Computer Science \\
  Johns Hopkins University  \\
  Baltimore, MD, USA \\
  \texttt{\{ssia1, aadelucia\}@jhu.edu}, 
  \hspace*{2em}
  \texttt{kevinduh@cs.jhu.edu} \\}

\date{}

\begin{document}
\maketitle
\begin{abstract}
Zero-shot In-context learning is the phenomenon where models can perform a task given only the instructions. However, pre-trained large language models are known to be poorly calibrated for zero-shot tasks. One of the most effective approaches to handling this bias is to adopt a contrastive decoding objective, which accounts for the prior probability of generating the next token by conditioning on a context. This work introduces an Anti-Language Model objective with a decay factor designed to address the weaknesses of In-context Machine Translation. We conduct our experiments across 3 model types and sizes, 3 language directions, and for both greedy decoding and beam search. The proposed method outperforms other state-of-the-art decoding objectives, with up to $20$ BLEU point improvement from the default objective in some settings. 

\vspace{1.5em}\hspace{.4em}\includegraphics[width=1.25em,height=1.25em]{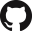}\hspace{.75em}\parbox{\dimexpr\linewidth-2\fboxsep-2\fboxrule}{\url{https://github.com/suzyahyah/icl_Anti-LM_decoding}}
\vspace{-.5em}
\end{abstract}

\section{Introduction}
Decoding strategies in supervised neural machine translation (NMT) models are typically designed to select the maximum likelihood response under the given model. However, in the era of in-context learning with large language models (LLMs), selecting for the maximum likelihood response should be re-examined, as LLMs are known to be poorly calibrated for their tasks and exhibit a strong prior bias \cite{zhao2021calibrate, wang2023reducing}. There are two large classes of decoding strategies: sampling methods and decoding objectives. Sampling methods do not change the probability ranking of the next token but influence how it is sampled, such as beam search \cite{koehn2004pharaoh, freitag-al-onaizan-2017-beam}, nucleus sampling \cite{Holtzman2020The}, and top-$k$ sampling \cite{fan-etal-2018-hierarchical}. Decoding objectives modify the probability of the next token before sampling takes place, typically by adding or subtracting scores, i.e., a \textit{contrastive objective}.

Decoding objectives are one of the most effective approaches for improving the output of a generative model. These often require no training and are typically “frustratingly simple”, because they only involve manipulation of output probability distributions at inference time. However, as outlined by \citet{zarriess2021decoding} in their recent large-scale survey, even for task-specific generation models, it is still surprisingly unclear what a good objective is for natural language generation.

In this work, we propose \textit{an anti-language model (Anti-LM) decoding objective with exponential decay}, which is motivated by the observation that LLMs are performing Bayesian inference under the hood \citep{xie2022an, mirchandani2023large}, and are thus inclined to continue generating in the source language. We investigate this under a zero-shot setting where only the instructions are provided to the model without any examples.\footnote{An example of the complete input to the model is ``\textit{Translate English to French: English: He built a WiFi door bell, he said. French:}''.}

We hypothesise that poor translations may be due to a strong prior bias of the dominant source language, but that this bias should diminish over future decoding steps. Anti-LM modifies the original logits by taking the difference of the next token logits, conditioned on the source sentence to be translated. Penalising the conditional source sentence logits discourages the model from continuing the non-translated generation from the source sentence or regurgitating it. We consider this negative scenario as a ``failure to translate''. 

Our work falls under the category of contrastive objectives, which were popularised by \citet{li-etal-2016-diversity}. We compare our approach against other contrastive objectives: ``conditional domain PMI'' \cite{holtzman-etal-2021-surface} and ``context-aware'' decoding \cite{shi2023trusting}. Our method consistently outperforms competitive baselines across language directions and model sizes, and the default objective by up to 20 BLEU points. Analysis shows that most of the gains come from the ``failure to translate'' case which the method was designed for. 

In addition, compared to the other contrastive decoding objectives, we only need to compute the contrastive logits (to be subtracted) once for the source sentence and not at every time step.  
\begin{table*}[!htbp]
    \centering
    \resizebox{2.1\columnwidth}{!}{
    \begin{tabular}{l|r|l|r}
    Name & RHS Expression & Example Conditional Text Input at \hl{$t=5$} & Note \\
    \midrule
    ALM$_u$ & $\gamma^{t} \log p(y_1 \mid  u)$ & \textcolor{blue}{Translate from English to French:} & Ablation with $u$ \\
    ALM$_x$ & $\gamma^{t} \log p(y_1 \mid  x)$ & \textcolor{red}{In summer, you'll need to watch out for mosquitoes.} & Our Method \\
    PMI$_u$ & $\alpha \log p(y_t| \mid  y_{<t}, u)$ & \textcolor{blue}{Translate from English to French:} \hl{En ete, il faudra} & \citet{holtzman-etal-2021-surface} \\
    PMI$_x$ & $\alpha \log p(y_t \mid  y_{<t}, x)$ & \textcolor{red}{In summer, you'll need to watch out for mosquitoes.} \hl{En ete, il faudra} & \citet{shi2023trusting} \\ 
    \bottomrule
    \end{tabular}}
    \caption{The four contrastive objectives evaluated in this work. The example shows the conditional input values for the following \textcolor{blue}{instruction ($u$)}, \textcolor{red}{source sentence ($x$)}, and model \hl{generation ($y_{<t}$)} at timestep $t=5$: \textit{\textcolor{blue}{Translate from English to French:} English: \textcolor{red}{In summer, you'll need to watch out for mosquitoes.} French:} \hl{En ete, il faudra}. PMI($u$) and PMI($x$) are shorthand for PMI($y; x \mid u$) and PMI$(y; u \mid x)$ respectively.}
    \label{tab:contrastive_exps}
\end{table*}


\section{Related Work and Background}
\paragraph{Zero-shot MT.} Prior work on zero-shot MT has focused on error analysis for the ``off-target'' problem \citep{tan-monz-2023-towards,chen-etal-2023-target},\footnote{Prior work defines ``off-target'' as the phenomena where the model returns a ``translation'' in the wrong language. In this work, we refer to both off-target and ``empty'' model generations as ``failure to translate''.} and techniques to improve translation performance \citep{gu-etal-2019-improved,chen-etal-2023-target,wen2024ebbs}.

Regarding error analysis, \citet{tan-monz-2023-towards} found that performance varies with respect to language direction, vocabulary overlap, and linguistic properties, and \citet{chen-etal-2023-target} identified lexical similarity between source and target language as an issue.
To combat the off-target problem, \citet{chen-etal-2023-target} introduced Language Aware Vocabulary Sharing (LAVS) to add language-specific tokens and decrease lexical similarity.
\citet{wen2024ebbs}, similar to our work, use a decoding  method to improve zero-shot performance. They introduced EBBS (Ensemble with Bi-level Beam Search), where multiple ``components'' influence the final generation. 
Other work uses a third language as a ``pivot'' \citep{gu-etal-2019-improved}.
Unlike these prior works, our method does not involve training new tokenizers, adding linear complexity with ensembling, or translating through other languages.

\paragraph{Contrastive Decoding} methods have been extensively explored for text generation, with different motivations behind each method. 
They have been used to reduce toxic language \citep{liu-etal-2021-dexperts}, improve general quality without further training \citep{li-etal-2023-contrastive}, improve factuality \citep{chuang2024dola}, and reduce repetition \citep{yang2023frustratingly}.  For summarization, \citet{van-der-poel-etal-2022-mutual} used conditional PMI decoding to avoid model hallucinations and promote ``faithfulness''.  
 \citet{shi2023trusting}, \citet{holtzman-etal-2021-surface} and \citet{kumar2022answer} adopt a weighted PMI based objective, conditioned on the context to ``penalise surface forms''.  The key difference between our Anti-LM formulation and other prior work, is that we compute the contrastive logits directly on $x$, the test source sentences to be translated, and not other ``non-$x$ context''. 
 
Our approach is motivated by improving the decoding of the target language by penalising source language continuations in Zero-shot MT. Within MT, concurrent work by \citet{sennrich-etal-2024-mitigating} also introduces a similar concept of ``language contrastive decoding'', but under a different formulation where they recompute the contrastive logits at each time step.  In contrast, our method does not require recomputing the logits at every time step, which greatly speeds up inference.

Similarly, sampling methods are also designed for different purposes. Nucleus sampling is good for creative generation \cite{delucia-etal-2021-decoding} while beam search is a popular choice for MT \cite{roberts2020decoding}. Decoding objectives can work in tandem with sampling methods but may have unexpected effects due to modification of the output probability space. We thus evaluate our objective with both Greedy Decoding and Beam Search.

\section{Method}

\subsection{Problem Formulation}
Let $x$ refer to the source test sentence, $y$ to the target test sentence to be generated, and $u$ to the instructions provided as context to the model. Autoregressive LMs generate text by a series of next-token predictions conditioned on the partial sequence generated so far. Greedy decoding proceeds by sampling the argmax token $y_t$ at every step $t$, given the previously sampled tokens $y_{<t}$, the test source sentence $x$ and the instructions $u$ based on the decoding objective $\log p(y_t | y_{<t}, x, u)$. \Cref{tab:contrastive_exps} summarises the evaluated contrastive objectives.

\subsection{PMI Decoding (Previous Work)}

An intuitive formulation of contrastive decoding is Pointwise Mutual Information ($\mathrm{PMI}$), where $\mathrm{PMI}(\mathbf{y}; \mathbf{x})$ measures the association between the target sequence $y$ and source sequence $x$. $\mathrm{PMI}(y; x)$ can be written as \footnote{Both $\log[\frac{p(y|x)}{p(y)}]$ and $\log[\frac{p(x|y)}{p(x)}]$ are equivalent forms. However $p(y|x)$ is more natural for autoregressive generation.} 

\begin{align*} 
\mathrm{PMI}(y; x) &= \log \frac{p(x, y)}{p(x) p(y)} \\ 
&= \log p(y|x) - \log p(y) 
\end{align*}

In $\mathrm{PMI}$ based objectives, the second term of \Cref{eq:anti-lm} functions as an anti-language model, and is typically weighted by $\alpha \in [0, 1]$.

\begin{equation}\label{eq:anti-lm}
    \hat{y} = \mathrm{argmax}_y \log p(y|x) - \alpha \log p(y)
\end{equation}

$\mathrm{PMI}$-based decoding (also known as Maximum Mutual Information Decoding \citet{li-etal-2016-diversity}) and its variants \citep{holtzman-etal-2021-surface, kumar2022answer, nandwani-etal-2023-pointwise} have been widely adopted in neural text generation. It penalizes high-frequency generic responses, but may also penalise fluent ones and thus can lead to ungrammatical outputs.

\paragraph{Conditional $\mathrm{PMI}$ Decoding}

$\mathrm{PMI}$ can also be interpreted as penalising the ``surface form'' \cite{holtzman-etal-2021-surface} of the target sequence, without having seen the source sequence in the context.

\begin{equation}\label{eq:pmi}
\log p(y_t | y_{<t}, x, u) - \alpha \log p(y_t | y_{<t}, u)
\end{equation}

The objective contains a penalty term for the log probability over the next token, conditioned on the target sequence decoded $y_{<t}$, and the context $u$. In our case the natural choice of $u$ would be the instructions \textit{``Translate <L1> to <L2>.''}.

\subsection{Anti-LM Contrastive Decoding}
We introduce our Anti-LM approach (ALM), which penalises the logits of the next token continuation of $x$, simply $\log p(y_1|x)$. 
The key difference between our Anti-LM objective and previous work is that we subtract logits conditioned directly on the test sentences $x$ to be translated, and not other contexts $u$ or any subsequent generations $y_{<t}$. Additionally, we use a discount factor $\gamma^t$ to reduce the influence of the Anti-LM on future timesteps. 

\begin{align}\label{eq:alm}
 \mathrm{ALM}(x) =& \log p(y_t|y_{<t}, x, u) 
 \\&-  \gamma^{t} \log p(y_1 | x) \nonumber
\end{align}

Unlike $\mathrm{PMI}$ decoding, the Anti-LM logits only need to be computed once for each source sentence. Note that $\log p(y_1|x)$ ensures that we never subtract the logits of the target language $y$ if there is a ``successful'' translation. As a control condition, we experiment with the Anti-LM conditioned on $u$, which has the same context as conditional $\mathrm{PMI}$.

\paragraph{Latency.}
Previous decoding methods require computation of the contrastive logits at every generation timestep, resulting in an additional time complexity of $O(n)$ where $n$ is the length of the string generated. In contrast the proposed method (regardless of choice of discount factor) is only $O(1)$ as it only needs to compute the contrastive objective once, and makes use of the decay factor.

\begin{table*}[]
\resizebox{\textwidth}{!}{
\begin{tabular}{llllllllllll}
\toprule
 &  & \multicolumn{5}{c}{\textbf{Greedy}} & \multicolumn{5}{c}{\textbf{Beam Search (B=5)}} \\
   \cmidrule(lr){3-7} \cmidrule(lr){8-12} 

  &  & base & $\text{PMI}_u$ & $\text{PMI}_x$ & $\text{ALM}_u$ & $\text{ALM}_x$ & base & $\text{PMI}_u$ & $\text{PMI}_x$ & $\text{ALM}_u$ & $\text{ALM}_x$  \\

\midrule
\multirow{6}{*}{en $\rightarrow$ fr} 
& XGLM 2.9B &    18.8 &  21.1 &  19.8 &      \underline{21.4} &      \textbf{21.5}  &       17.9 &   13.5 &  13.7 &      24.3 &      \textbf{26.6} \\
& XGLM 7.5B &    21.7 &  \underline{25.9} &  25.3 &      25.5 &      \textbf{26.0} &       13.2 &   20.9 &  17.4 &      28.1 &      \textbf{28.2} \\
& Bloom 3B  &    28.1 &  29.2 &  29.7 &      28.4 &      \textbf{30.0} &       30.4 &  28.3 &  30.5 &      30.2 &      \textbf{34.0} \\
& Bloom 7B &    32.5 &  32.7 &  33.3 &      32.7 &      \textbf{33.9}  &       34.6 &  32.8 &  32.9 &      34.8 &      \textbf{37.3} \\
& Llama 7B &  37.2 &  36.6 &  \textbf{37.3} &      37.0 &      36.6 &       38.1 &  35.4 &  34.4 &      \underline{38.6} &      \textbf{38.7}    \\
& Llama-chat 7B &    33.9 &  33.7 &  34.0 &      \underline{34.2} &      \textbf{34.3} &       34.6 &  32.6 &  32.6 &      34.9 &      \textbf{35.2} \\ 
\midrule
\multirow{6}{*}{en $\rightarrow$ pt} 
& XGLM 2.9B &     9.0 &  14.7 &  12.9 &      18.2 &      \textbf{19.6} &        5.9 &   7.4 &   14.8 &      19.8 &   \textbf{23.2}   \\
& XGLM 7.5B &    14.4 &  24.1 &  21.4 &      \textbf{25.4}  &      24.7  &        3.7 &  14.2 &  10.2  &      26.8 &      \textbf{27.0} \\
& Bloom 3b  &    29.9 &  30.3 &  \textbf{30.7}  &      30.0 &      \underline{30.6} &       32.1 &  30.3 &  31.3 &      31.7 &      \textbf{33.6} \\
& Bloom 7B &    32.1 &  \textbf{33.0} &  \underline{32.8} &      \textbf{33.0}  &      \underline{32.8}  &       35.6 &  34.0 &   33.7 &      \underline{35.7} &      \textbf{35.8}   \\
& Llama 7B  &    35.7 &  35.5 &  \textbf{35.9} &      35.4 &      35.6 &       36.9 &  35.2 &  34.7 &      36.7 &      \textbf{37.4} \\

& Llama-chat 7B &    32.9 &  33.0 &  33.0 &      33.2 &      \textbf{33.4} &       34.0 &  31.9 &  31.7 &      \textbf{34.4} &      \textbf{34.4}  \\
  \midrule
\multirow{6}{*}{en $\rightarrow$ de} 
& XGLM 2.9B &    12.0 &  \textbf{13.6}  &  12.7 &      13.2 &      13.3  &       11.9 &   8.9 &   8.4  &      16.0 &      \textbf{17.6} \\
& XGLM 7.5B &    11.7 &  16.3 &  15.0 &      17.5 &      \textbf{17.8}  &        4.1 &  10.8 &   7.9 &      18.2 &      \textbf{18.5} \\
& Bloom 3b  &     3.3 &   3.9 &   3.6 &       3.8 &       \textbf{4.6}  &        3.5 &   3.8 &   3.7 &       3.7 &       \textbf{5.0} \\
& Bloom 7B &     3.1 &   \textbf{8.2} & \underline{8.0}   &       7.9 &       \underline{8.0}  &        7.8 &   8.8. & 7.4 &       8.1 &       \textbf{9.0} \\
& Llama 7B &   \underline{25.5} &  25.1 &  \textbf{25.6} &      25.3 &      \underline{25.5} &       25.5 &  24.7 &  23.8 &      26.0 &     \textbf{27.1} \\
&  Llama-chat 7B &  22.5 &  22.3 &  22.5 &      22.7 &      \textbf{23.2} &       \underline{23.5} &  21.6 &  21.2 &      \textbf{23.7} &      23.4  \\
\bottomrule
\end{tabular}}
\caption{Translation performance on FLORES with greedy decoding and beam search ($B=5$). Scores are reported with SacreBLEU \citep{post-2018-call}, where higher is better. ``base" refers to default maximum likelihood decoding. The best scores are bolded and scores within 0.2 of the best are underlined. \{\} $\rightarrow$ en results are in Appendix \Cref{fig:logits_x_en}.}
\label{tab:results_baseline}
\end{table*}

\section{Experiments}
\label{sec:experiments}

\paragraph{Decoding Objectives} We evaluate 4 decoding objectives in addition to the default maximum likelihood objective (summarised in \Cref{tab:contrastive_exps}).

\paragraph{Models.}
\label{sec:models}
We use three models: XGLM (2.9B, 7.5B) \citep{lin-etal-2022-shot}, Bloom (3B, 7B) \citep{scao2022bloom} and Llama 2 (7B, 7B-chat) \citep{touvron2023llama2}.\footnote{We also experimented with OPT2.7B \citep{zhang2022opt} but found that its in-context MT abilities were very poor. The RLHF version of Bloom, Bloomz7B reached a suspiciously high BLEU score of 60 and we suspect data leakage during its training.} All models are available on HuggingFace \citep{wolf-etal-2020-transformers}, with the latter three having been advertised as ``Multilingual Language Models.'' 
To our knowledge, there have not been any reports of sentence-level parallel corpora in their training datasets (\Cref{sec:appendix_models}). In other words, these models were not trained with data that explicitly supports the translation task. 

\paragraph{Data and Evaluation.}
We evaluate on the Wikipedia-based FLORES-101 \citep{goyal-etal-2022-flores} in three bi-directions with English: French (en$\leftrightarrow$fr), German (en$\leftrightarrow$de), and Portuguese (en$\leftrightarrow$pt). As these are zero-shot experiments, no separate dataset is required to be used as the prompt bank, and no randomness is associated with prompt selection. 

For evaluation, we report BLEU \citep{papineni-etal-2002-bleu} and COMET-22 \citep{rei-etal-2022-comet}, two reference-based automatic metrics. COMET-22 is a neural-based method reported to correlate highly with human judgment \citep{freitag-etal-2022-results}. We use the SacreBLEU \citep{post-2018-call} implementation of BLEU with default arguments and Unibabel's COMET-22 implementation.\footnote{\url{https://github.com/Unbabel/COMET}}

\paragraph{Generation Settings.}
For the decoding objectives (\Cref{tab:contrastive_exps}), we chose $\alpha=0.1$ for PMI Decoding and $\gamma=0.3$ for Anti-LM decoding (see \Cref{sec:hyperparam}) after hyperparameter search on only a single language direction (en $\rightarrow$ pt) using the dev set, thereafter applying the same $\alpha$ to all experiments.
We evaluate on both greedy decoding and beam search (B=$5$).

\paragraph{Instructions.}
We provide instructions in the source (L1) language using the instructions "Translate from <L1> to <L2>" and the "masterful translator" prompt by \cite{reynolds2021prompt}. See Appendix \Cref{tab:prompts} for details. 

\section{Results}
We observe that \textbf{Anti-LM objective is best across most objectives, language directions, and sampling strategies} (see \Cref{tab:results_baseline}), although this is less pronounced in Llama7B. We find that PMI outperforms the default objective, which is consistent with previously reported work.
For beam search, the Anti-LM objective is particularly effective for XGLM with an improvement of BLEU by up to 20 points.
Example translations and COMET scores are in \Cref{sec:appendix_failure}.

\section{Analysis}
\label{sec:analysis}

\begin{figure}[!t]
    \centering
\includegraphics[width=\columnwidth]{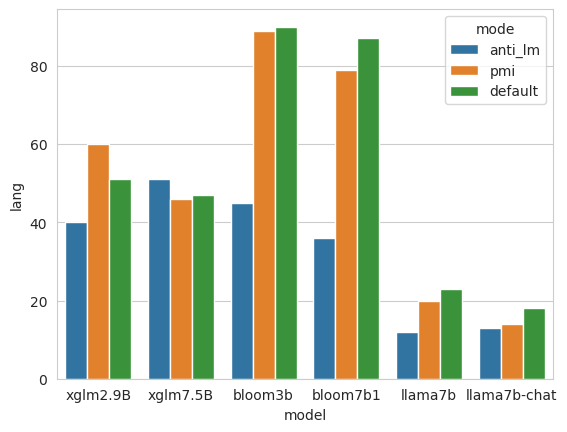}
\caption{Number of non-target French sentences generated given the task \textit{Translate English to French} which indicates a failure to translate.}
    \label{fig:translate_fail_fr}
\end{figure}

\paragraph{Failure to Translate.}
Models may fail to translate the provided sentence due to no generation or generation in the source (L1) language. Even for the ``large'' multilingual models (XGLM7.5B and Bloom7B), the models still make a sizeable number of such errors (10\%-45\%). 
\Cref{fig:translate_fail_fr} shows the number of translation failures across models for $\mathrm{PMI}(x)$ and Anti-LM($x$) for en$\leftrightarrow$fr against the default (greedy) objective. 

\begin{figure}[!t]
    \centering
\includegraphics[width=\columnwidth]{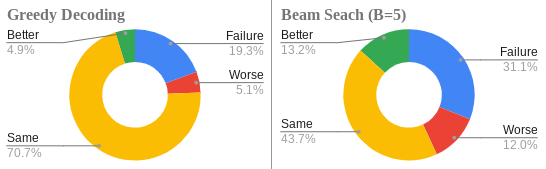}
\caption{Proportion of failure to translate vs successful cases averaged across all models. For successful cases, we compute whether the Anti-LM objective improves, degrades, or does not affect the performance.}
\label{fig:doughnut}
\end{figure}

We analyse the scores for the non-failure cases and find that there is largely equivalent proportion of sentences which are either better or worse than the baseline (\Cref{fig:doughnut}). 
This indicates that the \textbf{gains observed can be attributed to addressing failure to translate cases} (see \Cref{sec:appendix_failure}).

\paragraph{Missing Entity Rate.}
One aspect of translation ``faithfulness'', considers the named entity retention from the source to the target \citep{alves-etal-2022-robust}. For example, the name ``Ehud Ur'' should be included as-is in the translation. This is one potential area for improvement of the proposed approach, as the contrastive objective would affect the logits of the "as-is" named entities (see \Cref{sec:mer}).

\paragraph{Choice of Discount Factor.}
The discount factor presented in \Cref{eq:alm} is an exponential decay in timesteps $t$. We investigated the importance of the decay function by evaluating others: reverse-ReLU, logistic, and Gompertz decay, which is an asymmetric logistic function. Details of these functions are in \Cref{sec:decay_functions}. We found that Gompertz decay and reverse-ReLU can sometimes outperform exponential, although their performances are quite similar (\autoref{tab:results_discount}). 
\begin{table}[]
\resizebox{\columnwidth }{!}{
\begin{tabular}{lllllll}

 \toprule
  &  & base & exp & r-relu & log & gomp  \\
\cmidrule(lr){3-7}

\multirow{6}{*}{\rotatebox[origin=c]{90}{en $\rightarrow$ fr}} 
& XGLM 2.9B &    19.1 &      \textbf{21.3} &           21.0 &               20.6 &    21.0 \\
& XGLM 7.5B &   21.7 &      26.0 &           \underline{26.2} &               25.2 &               \textbf{26.4} \\
& Bloom 3B  & 28.2 &      \textbf{29.9} &           \underline{29.8} &               29.1 &               \textbf{29.9} \\
& Bloom 7B &   32.4 &      \textbf{33.8} &           \underline{33.6} &               32.9 &               \underline{33.7} \\
& Llama 7B &  \textbf{37.2} &      36.6 &           36.1 &               35.7 &               36.3 \\
& Llama-chat 7B & 33.9 &      34.3 &           34.5 &               \textbf{34.8} &               34.4 \\
\midrule
\multirow{6}{*}{\rotatebox[origin=c]{90}{en $\rightarrow$ pt}} 
& XGLM 2.9B     &     8.9 &      19.3 &           \textbf{20.0} &               19.0 &               19.6 \\
& XGLM 7.5B     &    14.3 &      25.1 &           \textbf{26.2} &               24.7 &               25.6 \\
    & Bloom 3B      &    29.9 &     \textbf{30.7} &           30.4 &               29.4 &               \underline{30.5} \\
& Bloom 7B     &    32.1 &      \textbf{32.9} &           32.7 &               31.9 &               \underline{32.8} \\
& Llama 7B      &    \textbf{35.7} &      \underline{35.6} &           \textbf{35.7} &               35.1 &               35.4 \\
& Llama-chat 7B &    32.9 &      \underline{33.4} &           \textbf{33.6} &               \underline{33.5} &               \textbf{33.6} \\
  \midrule
\multirow{6}{*}{\rotatebox[origin=c]{90}{en $\rightarrow$ de}} 
& XGLM 2.9B     &    12.0 &      \textbf{13.6} &           13.3 &               12.5 &               \textbf{13.6} \\
& XGLM 7.5B     &    11.8 &      17.7 &           17.5 &               16.8 &               \textbf{17.9} \\
& Bloom 3B      &     3.3 &       \textbf{4.5} &            \textbf{4.5} &                4.3 &                \textbf{4.5} \\
& Bloom 7B     &     7.3 &       \textbf{8.0} &            7.6 &                7.5 &                7.8 \\
& Llama 7B      &    \textbf{25.5} &      \textbf{25.5} &           \underline{25.3} &               23.7 &               \underline{25.3} \\
& Llama-chat 7B &    22.5 &      \textbf{23.2} &           \textbf{23.2} &               23.1 &               23.1 \\

\bottomrule
\end{tabular}}
\caption{Translation performance on FLORES with greedy decoding using different decay functions. These are the exponential (exp) as shown in eq 3, reverse-relu (relu), logistic (log), and gompertz decay (gomp). The best scores are bolded and scores within 0.2 of the best are underlined.}
\label{tab:results_discount}
\end{table}

\paragraph{Instruction Language.}
Anti-LM has a positive effect in the L1$\rightarrow$ L2 direction, \textit{if} the instructions were given in the L1 (source) language. Our findings indicate that there is an unintended effect of source language dominance during zero-shot MT.\footnote{We find that the approach is also effective for translating from other languages, e.g., French to Portuguese.} This suggests that without taking into account the Anti-LM calibration, the \textit{true} zero-shot capabilities of GPT-style models may be under-reported.

\paragraph{Elaborate Instructions.}
Anti-LM similarly outperforms the baseline and comparisons in an experiment with more elaborate instructions, specifically the ``masterful translator'' prompt by \citet{reynolds2021prompt}. See \Cref{app:alt_instruct} and \Cref{tab:prompts} for details and results.
\section{Conclusion}
Decoding objectives are one of the most effective ways to improve a model's output, especially if it has strong prior bias from pre-training. We designed an Anti-LM objective with decay for zero-shot Machine Translation which has a much smaller computational overhead and is more effective than existing approaches. Our method outperforms strong baselines across language directions, model types and sizes, and decoding strategies, especially in failure to translate cases.

\clearpage
\section{Limitations}
\paragraph{Comparison to Few-shot.}
The approach described in this work while effective in the zero-shot setting was found to be less effective for K-shot examples setting. We did not tune the hyperparameter for this setting or investigate this thoroughly. The primary reason is that the K-shot examples has much less failure to translate cases, i.e., more consistent at giving an appropriate translation in the target language. 

\paragraph{Low-resource Languages.} We do not evaluate our method on 'low-resource' languages. However, what the MT community traditionally considers as `low-resource' language is a misnomer when working with pre-trained language models, as a language might be `low-resource' for the model if it is not explicitly collected in the training data. An example of this is German (de) for Bloom. While traditionally considered a `high-resource' language, it is actually a `low-resource' language for Bloom as it was not collected in the dataset.\footnote{\url{https://huggingface.co/bigscience/bloom/discussions/221}}

\paragraph{Human Evaluation.}
While we do evaluate with COMET-22, a metric well-correlated with human judgment, we did not include a human annotation study for the generations.

\section{Ethics Statement}

In the course of LLM generation, there may be unexpected outputs. The generations of our method may have hallucinated content and can be misleading. When deployed in real-world applications, special attention should be paid to avoid inappropriate generations. For example, one can use post-process steps such as fact-checking for named entities. With regard to toxic or unfair output, we believe that the method does not contribute to these biases that were not already previously present in the pre-trained models.

\bibliography{anthology,custom}
\bibliographystyle{acl_natbib}

\clearpage
\appendix
\section{Models}
\label{sec:appendix_models}
We ran all experiments on 24GB GeForce RTX 3090 and 32GB Tesla V100 GPUs for the models\footnote{In an earlier version of this paper, we had included GPTNeo results which show large positive effects in the greedy decoding case, and mixed results in the Beam Search case. We omit GPTNeo in this version due to space.} under and over 7B parameters, respectively.

XGLM adopts a similar architecture to GPT-3 \citep{brown2020language} and was trained on the large multilingual Common Crawl (CC100-XL, \citep{conneau-etal-2020-unsupervised}). Bloom has been trained on the ROOTS Corpus \citep{laurenccon2022bigscience}, a 1.6 TB collection of HuggingFace text datasets. 
Llama 2 was trained on an unspecified ``new mix of publicly
available online data'', however, it is 90\% English \citep{touvron2023llama2}. 
This mix most likely includes the Llama training set from CommonCrawl, C4, GitHub, Wikipedia, Gutenburg, Books3, ArXiv, and StackExchange, some of which are multilingual \citep{touvron2023llama}.

\section{Hyperparameter Selection}
\label{sec:hyperparam}
\subsection{Hyperparameter Sweep}
\begin{figure}[!h]
    \centering
    \includegraphics[width=\columnwidth]{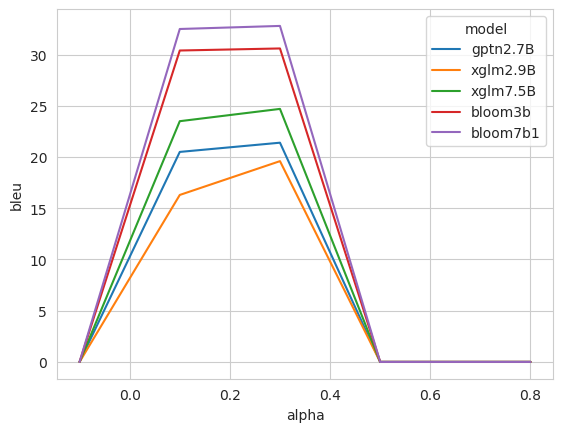}
    \caption{$\gamma$ sweep for $\texttt{en} \rightarrow \texttt{pt}$.}
    \label{fig:hyperparam_sweep}
\end{figure}

The only hyperparameters associated with our experiments are $\alpha$ on PMI Decoding and $\gamma$ for the discount factor on Anti-LM Decoding. We experiment with $\{-0.1, 0.1, 0.3, 0.5, 0.8, 1.0\}$ for both $\alpha$ and $\gamma$ and observe that the best $\alpha$ is $0.1$ and the best $\gamma$ is $0.3$ across models. \Cref{fig:hyperparam_sweep} is an example graph of the hyperparameter sweep for $\gamma$. 

Note that we only search for the hyper-parameter once on the \textbf{dev set} of en $\rightarrow$ pt, and use the same hyper-parameter throughout all experiments. i.e., We did not tune the hyperparameter for every single language direction. Note also that the hyperparameters found generalises \textit{across} models. We do not perform hyper-parameter search with Llama models and adopt the same hyperparameter that was found with other models.

\begin{table*}[]
\resizebox{\textwidth}{!}{
\begin{tabular}{llllllllllll}
\toprule
 &  & \multicolumn{5}{c}{\textbf{Greedy}} & \multicolumn{5}{c}{\textbf{Beam Search (B=5)}} \\
   \cmidrule(lr){3-7} \cmidrule(lr){8-12} 

  &  & base & $\text{PMI}_u$ & $\text{PMI}_x$ & $\text{ALM}_u$ & $\text{ALM}_x$ & base & $\text{PMI}_u$ & $\text{PMI}_x$ & $\text{ALM}_u$ & $\text{ALM}_x$  \\

\midrule
\multirow{6}{*}{fr $\rightarrow$ en} & Xglm 2.9B & 23.3 &  29.1 &  27.4 &      \textbf{30.1} &      \underline{30.0} &       19.8 &  23.1 &  20.5 &      \textbf{31.4} &      \underline{31.3}  \\
& Xglm 7.5B & 24.6 &  33.1 &  31.9 &      34.4 &      \textbf{34.8} &       12.0 &  31.2 &  26.8 &      34.8 &      \textbf{35.2} \\
& Bloom 3B & 14.0 &  27.6 &  22.0 &      29.9 &      \textbf{32.7} &       26.3 &  31.6 &  28.2 &      27.0 &      \textbf{34.8} \\ 
& Bloom 7B & 24.7 &  31.9 &  29.7 &      32.0 &      \textbf{37.2} &       25.7 &  36.6 &  34.1 &      27.3 &      \textbf{37.6}\\
& Llama7b & 40.7 &  \textbf{41.4} &  40.8 &      40.8 &      41.0 &       40.1 &  39.6 &  39.6 &      40.0 &      \textbf{40.4}  \\
& Llama7bc & \underline{40.3} &  \textbf{40.5} &  40.2 &      40.2 &      40.0 &       \textbf{40.3} &   40.2 &   \underline{40.1} &      \textbf{40.3} &      \underline{40.1}\\
\midrule
\multirow{6}{*}{pt $\rightarrow$ en} & Xglm 2.9B & 29.7 &  32.3 &  32.1 &      \underline{33.2} &      \textbf{33.3} &       27.1 &  30.3 &  29.2 &      \textbf{35.2} &      34.6 \\ 
& Xglm 7.5B & 18.7 &  \textbf{38.3} &  37.6 &      37.8 &      \underline{38.2} &        7.8 &  31.8 &  28.8 &      \textbf{39.4} &      39.0\\
& Bloom 3B & 19.1 &  34.9 &  33.3 &      28.9 &      \textbf{35.2} &      25.1 &  34.0 &  31.4 &      34.9 &      \textbf{38.0} \\
& Bloom 7B & 27.5 &  36.0 &  35.5 &      15.0 &      \textbf{37.4} &    29.5 &  35.0 &  33.3 &      29.3 &      \textbf{39.6} \\
& Llama 7B & 38.4 &  \textbf{44.0} &  41.2 &      42.8 &      \textbf{44.0} &     42.0 &  41.6 &  35.1 &      42.8 &      \textbf{43.7} \\
& Llama-chat 7B & 43.2 &  \textbf{43.6} &  42.8 &      43.2 &      43.2 &       \underline{43.2} &  \textbf{43.4} &   43.1 &      \underline{43.2} &      \underline{43.3} \\
\midrule

\multirow{6}{*}{de $\rightarrow$ en} & Xglm 2.9B & 5.3 &   4.7 &   5.1 &       8.8 &       7.2 &        2.0 &   3.8 &   3.3 &       2.2 &       2.2 \\
& Xglm 7.5B & 32.3 &  32.3 &  32.2 &      32.5 &      \textbf{33.0} &       29.3 &  31.0 &  29.3 &      33.2 &      \textbf{33.8} \\
& Bloom 3B & 7.1 &   7.2 &   7.4 &       7.9 &      \textbf{ 9.1} &        6.4 &   6.2 &   6.3 &       6.7 &       \textbf{7.6}\\
& Bloom 7B & 20.0 &  18.8 &  20.2 &      20.2 &      \textbf{21.1} &       18.0 &  18.5 &  19.8 &      18.4 &      \textbf{19.5} \\
& Llama 7b & 39.3 &  40.4 &  39.7 &      \underline{39.4} &      \textbf{39.5} &       37.6 &  37.9 &  36.7 &      \underline{38.4} &      \textbf{38.5} \\
& Llama 7b-chat & \textbf{39.3} &  \underline{39.2} &  39.0 &      \underline{39.1} &      \underline{39.1} &       38.9 &  \textbf{39.2} &   \underline{39.0} &      \underline{39.0} &      38.9 \\
\bottomrule
\end{tabular}}
\caption{Translation performance on FLORES with greedy decoding and beam search ($B=5$). Scores are reported with SacreBLEU \citep{post-2018-call}, where higher is better. ``base" refers to default maximum likelihood decoding. The best scores are bolded and scores within 0.2 of the best are underlined.}
\label{fig:logits_x_en}
\end{table*}

\section{COMET Results}
\label{sec:comet_results}
The COMET results are shown in \Cref{tab:results-comet}. The COMET score ranges from 0 to 1, where a score of 1 is considered a good translation. While the scores appear high, they should be interpreted as a comparative score instead of an absolute score. 

Improvements over the baselines are primarily seen with greedy decoding. And as in \Cref{tab:results_baseline}, XGLM benefits the most from the calibration offered from contrastive decoding. Also, the ALM objectives more consistently improved over the baselines than the PMIs.

\paragraph{Comparing BLEU and COMET scores}
For a fair comparison to how we evaluated with BLEU, we did not remove the non-L2 generations. Unlike BLEU, COMET still awards higher than expected scores to ``translations" that should be considered failure cases. For example, a model can get scores of $70$ and $34$ for simply repeating the source sentence (i.e., generating L1 instead of L2) and not generating anything, respectively. These failures are further discussed in \Cref{sec:appendix_failure}.

\section{Failure Analysis}
\label{sec:appendix_failure}
We evaluate ``failure to translate" in multiple ways. As discussed in \Cref{sec:analysis}, the most common failure case is when the model generates the L1 (source) language instead of L2 (target). Statistics on how often that occurs across all the methods and models are shown in \Cref{tab:failure_comparison}.

\subsection{Rate of Empty Generation (REG)}

Separate from the L1 generation is when the model does not generate a response at all. We refer to this as ``empty generation", and the Rate of Empty Generation (REG) is shown in \Cref{tab:results-REG}. Since this measurement is the ratio of the number of empty generations to the number of generations, a score of $0$ is best and a score of $100$ is very poor. Though generating text does not mean it is correct or in L2, only that there was \textit{some} output from the model.

An interesting note is that the REG of the baselines (i.e., without special decoding objectives) are never $0$, which occurs more frequently with PMI and ALM objectives. Regardless of the decoding method, the ALM(u) objective has the best REG. From the ALM(u) and PMI(u) scores, it is apparent that conditioning on the instructions (u) reduces the number of empty generations. Overall, greedy decoding produces the highest rate of output from the models.

\subsection{Missing Entity Rate (MER)}
\label{sec:mer}
Another failure case, which can be used as a proxy for translation ``faithfulness", considers the named entity retention from the source to the target. For example, a name such as ``Ehud Ur" should be included as-is in the translation (see \Cref{tab:failure_ex}). We define the Missing Entity Rate (MER) as the ratio of the number of entities that \textit{should} be in the translation (as determined by the reference) out of all entities in the source. We use the \texttt{en\_core\_web\_trf} spaCy model to extract entities from the source sentences \citep{spacy}. The model is penalised for not generating any text, and only source sentences that have at least one detected entity are considered. This metric is similar in spirit to the ``deviation in named entities" challenge from SMAUG \citep{alves-etal-2022-robust}. Similar to the trend with REG scores, we see that the contrastive objectives outperform the baselines (\Cref{tab:results-MER}), and the improvements are greater when conditioning on the instructions ($u$).

\section{Decay Functions}
\label{sec:decay_functions}
\begin{figure}[!h]
    \centering
    \includegraphics[width=\columnwidth]{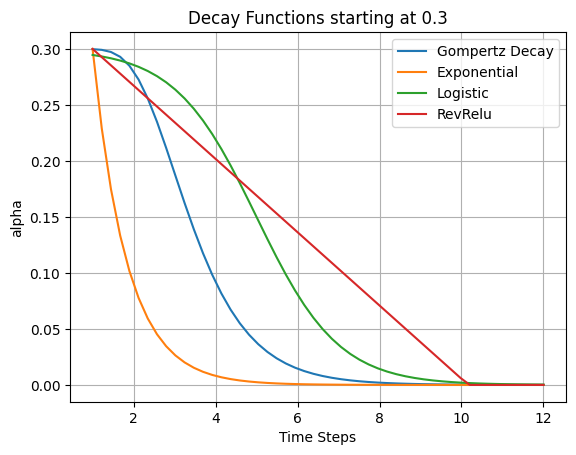}
    \caption{Weighting by different options for the decay functions, starting at 0.3 and ending at 0 in the \Cref{eq:anti-lm}. Aside from exponential, all decay functions have the additional parameter of when the function reaches 0, we set this to 10 timesteps.}
    \label{fig:decay_functions}
\end{figure}

The following decay functions were evaluated for the analysis in \Cref{{tab:results_discount}}. The shape of these functions are shown in \Cref{fig:decay_functions}.

\begin{itemize}
    \item \textbf{Gompertz Decay} with parameters $a=0.3, b=20, c=1$.  
    $$f(t) = a * \textrm{exp}( -b * \textrm{exp}(-c * t))$$
    \item \textbf{Exponential} with $a=0.3$
    $$f(t) = a^t$$
    
    \item \textbf{Logistic} with $a=0.3, k=1, t0=5$.
    $$f(t) = -a / (1 + \textrm{exp}(-k*(t - t0))) + a$$
    \item \textbf{Rev-Relu}\footnote{This is not an official name.} with $a=0.3$.
    $$f(t) = max(0, -a*t + a)$$

\end{itemize}

\section{Alternative Instructions}
\label{app:alt_instruct}

To ensure the results are not prompt-specific, we also ablate with another translation task prompt, the ``masterful translator'' from \citet{reynolds2021prompt}: \textit{“A <L1> phrase is provided. The masterful <L1> translator flawlessly translates the phrase into <L2>:”} (see \Cref{tab:prompts}). 
This prompt is more specific than the original, which is expected to cause the model to generate a response with the intended behavior, in this case, an accurate translation.
Overall, we observed that while the model improved with a better prompt, the improvement of prompt-tuning \textit{and} ALM decoding was greater.
This indicates that zero-shot models can benefit from both prompt-tuning and decoding methods.
See below for details for each metric.

\paragraph{BLEU}
From \Cref{tab:results_masterful}, we observe that while the individual scores for each model are higher for over half the models (64\%, designated in the table with italics), the same trends as with the basic prompt exist; the ALM decoding methods improve over the baseline in almost all cases and decoding methods.
The average ALM($x$) improvement over the baseline is higher with the masterful prompt than with the basic prompt, with average gains of +2.52 with greedy search (compared to +0.55) and +5.53 with beam search (compared to +4.77).

\paragraph{COMET}
The COMET results follow the same trend as the BLEU results, with improving the model performance 58\% of the time. 
The average ALM($x$) improvement over the baseline is higher with the masterful prompt than with the basic prompt, with average gains of +4.10 with greedy search (compared to +3.61) and +7.75 with beam search (compared to +6.80).

\paragraph{Missing Entity Rate (MER)}
The improvements of the masterful prompt over MER are less pronounced as with BLEU and COMET.
The model performance is only improved 41\% of all cases.
The average ALM($x$) improvement over the baseline is higher with the masterful prompt than with the basic prompt, with average gains of -5.85 with greedy search (compared to -1.12) and -12.04 with beam search (compared to -11.42).

\paragraph{Rate of Empty Generation (REG)}
The REG improvements are more pronounced than with MER but still less than BLEU and COMET, with increased model performance 45\% of the time. 
The average ALM($x$) improvement over the baseline is higher with the masterful prompt than with the basic prompt, with average gains of -8.44 with greedy search (compared to -2.78) and -14.03 with beam search (compared to -13.67).

\newcommand{\source}{\texttt{\{SOURCE\}}\xspace}
\newcommand{\lone}{\texttt{\{L1\}}\xspace}
\newcommand{\ltwo}{\texttt{\{L2\}}\xspace}

\begin{table*}[]
    \centering
    \begin{tabular}{lll}
    \toprule
    \multirow[c]{4}{*}{Prompt} & (en) & Translate \lone to \ltwo: \lone: \source \ltwo: \\
     & (de) & \resizecol{0.65}{Übersetzen Sie vom Deutschen ins Englische: Deutschen: \source Englische:}\\
     & (fr) & \resizecol{0.65}{Traduire du français vers l'anglais: français: \source l'anglais:}\\
     & (pt) & \resizecol{0.65}{Traduzir português para inglês: português: \source inglês:}\\
    \midrule
    ``Masterful'' Prompt & & \resizecol{0.65}{A \lone phrase is provided. The masterful \lone translator flawlessly translates the phrase into \ltwo: \lone: \source \ltwo:} \\
    \bottomrule
    \end{tabular}
    \caption{The two prompt templates used in the experiments. \lone refers to the source language and \ltwo refers to the target language. \source is replaced with the source-language sentence for translation. The prompt is translated into the source language for German (de), French (fr), and Portuguese (pt). The ``masterful'' prompt is from \citep{reynolds2021prompt}.}
    \label{tab:prompts}
\end{table*}

\begin{table*}[!h]
\centering
\begin{tabular}{llrrrr}
\toprule
 &  &  \multicolumn{3}{c}{Non-failure} &  Failures \\
Sampling & Model&  Better &  Equal &  Worse &   \\
\midrule
Default & XGLM2.9B &       0.12 &       0.78 &       0.12 &    0.30 \\
& XGLM7.5B &  0.08 &       0.83 &       0.08 &    0.33 \\
& Bloom3B &       0.07 &       0.88 &       0.05 &    0.30 \\ 
& Bloom7B & 0.05 &       0.90 &       0.05 &    0.13 \\
& Llama7B &  0.08 &       0.87 &       0.08 &    0.03 \\
& Llama7B-chat &  0.00 &       0.97 &       0.00 &    0.07 \\
\midrule
Beam Search & XGLM2.9B &   0.22 &       0.52 &       0.23 &    0.43 \\
& XGLM7.5B &  0.20 &       0.57 &       0.22 &    0.62 \\
& Bloom3B &    0.17 &       0.62 &       0.20 &    0.40 \\ 
& Bloom7B &     0.17 &       0.62 &       0.20 &    0.22  \\
& Llama7B &    0.20 &       0.53 &       0.22 &    0.10\\
& Llama7B-chat &  0.10 &       0.80 &       0.10 &    0.07 \\
\bottomrule
\end{tabular}
\caption{Proportion of better, equal or worse scoring sentences where the difference is at least 5 BLEU points, when comparing the AntiLM approach against the baseline, when excluding `failure to translate' cases. All values are aggregated across three language directions, en $\rightarrow$ {fr, pt, de}.} 
\label{tab:failure_comparison}
\end{table*}

\begin{table*}[]
\centering
\resizebox{!}{4.21in}{\begin{tabular}{cclll}
\multicolumn{4}{l}{Source: He built a WiFi door bell, he said.} \\
\multicolumn{4}{l}{Target: Il dit avoir conçu une sonnette de porte Wi-Fi.} \\
\toprule

\multicolumn{1}{l}{} & \multicolumn{1}{l}{} &  & \multicolumn{2}{c}{Translation} \\
\multicolumn{1}{l}{} & Objective & \multicolumn{1}{c}{Model} & \multicolumn{1}{c}{Regular Prompt} & \multicolumn{1}{c}{Masterful Prompt} \\ \midrule

\parbox[t]{5mm}{\multirow{30}{*}{\rotatebox[origin=c]{90}{Beam Search (B=5)}}} & \multirow{6}{*}{ALM(u)} & Bloom3B & Il construisit un interphone sans fil, il a dit. & Il construisit une sonnette WiFi, il a dit. \\
 &  & Bloom7B & Il a construit un interphone sans fil, il a dit. & Il a construit un interphone sans fil, il a dit. \\
 &  & Llama2-7B & Il a construit un timbre à sonner par WiFi, il a dit. & Il a construit une sonnette WiFi, a-t-il dit. \\
 &  & Llama2-7BChat & Il a construit un timbre Wi-Fi. & Il a construit un appareil de sonnette Wi-Fi, il a dit. \\
 &  & XGLM2.9B & Il a construit un WiFi, il a dit. & Il a construit une alarme WiFi, il a dit. \\
 &  & XGLM7.5B & Il a construit un interphone WiFi, il a dit. & He a construit un WiFi porte-clefs, il a dit. \\
  \cline{2-5}
 & \multirow{6}{*}{ALM(x)} & Bloom3B & Il a construit un sonnette WiFi, il a dit. & Il construisit un sonnette WiFi, il a dit. \\
 &  & Bloom7B & Il a construit un interphone sans fil, il a dit. & Il a construit un interphone sans fil, il a dit. \\
 &  & Llama2-7B & Il a construit un appareil de sonnette Wi-Fi, il a dit. & Il a construit une sonnette Wi-Fi, a-t-il dit. \\
 &  & Llama2-7BChat & Il a construit un timbre Wi-Fi. & Il a construit un appareil de sonnette Wi-Fi, il a dit. \\
 &  & XGLM2.9B & Il a construit une WiFi porte-clef, il a dit. & Il a construit une WiFi porte-clés, il a dit. \\
 &  & XGLM7.5B & Je l'ai construit un WiFi porte-clefs, il a dit. & Il a construit un interphone sans fil, il a dit. \\
  \cline{2-5}
 & \multirow{6}{*}{PMI(u)} & Bloom3B & Il a construit un buzzer WiFi, dit-il. & Il a construit un interphone WiFi, il a dit. \\
 &  & Bloom7B & Il a construit un bouton d'appel WiFi, il a dit. & Il a construit une sonnette WiFi, il a dit. \\
 &  & Llama2-7B & Il a construit un interphone WiFi, il a dit. & Il a construit une sonnette WiFi, il a dit. \\
 &  & Llama2-7BChat & Il a construit un timbre WiFi. & Il a construit un interphone Wi-Fi, il a dit. \\
 &  & XGLM2.9B & Il a construit un WiFi interphone. & Il a construit un WiFi interphone. \\
 &  & XGLM7.5B & \textless{}EMPTY GENERATION\textgreater{} & Il a construit un interphone WiFi, il dit. \\
  \cline{2-5}
 & \multirow{6}{*}{PMI(x)} & Bloom3B & Il a construit un porte-clés WiFi, dit-il. & Il a construit un petit interphone sans fil, il a dit. \\
 &  & Bloom7B & Il a construit une sonnette WiFi. & Il a construit un interphone sans fil, il a dit. \\
 &  & Llama2-7B & Il a construit un appareil de sonnette WiFi, il a dit. & Il a construit une sonnette WiFi, il a dit. \\
 &  & Llama2-7BChat & Il a construit un timbre WiFi. & Il a construit un appareil de sonnette Wi-Fi, il a dit. \\
 &  & XGLM2.9B & Il a construit une wifi porte-fenêtre. & Il a construit une alarme WiFi. \\
 &  & XGLM7.5B & \textless{}EMPTY GENERATION\textgreater{} & \textless{}EMPTY GENERATION\textgreater{} \\
  \cline{2-5}
 & \multirow{6}{*}{base} & Bloom3B & Il a construit un sonnette WiFi, il a dit. & Il a construit une sonnette WiFi, il a dit. \\
 &  & Bloom7B & Il a construit un interphone sans fil, il a dit. & Il a construit une sonnette WiFi, il a dit. \\
 &  & Llama2-7B & Il a construit un appareil de sonnette WiFi, il a dit. & Il a construit une sonnette WiFi, a-t-il dit. \\
 &  & Llama2-7BChat & Il a construit un timbre Wi-Fi. & Il a construit un interphone Wi-Fi, il a dit. \\
 &  & XGLM2.9B & Il a construit un WiFi, il a dit. & Il a construit un WiFi. \\
 &  & XGLM7.5B & Il a construit un interphone WiFi, il a dit. & Il a construit un WiFi porte-clefs, il a dit. \\
 
 \midrule\midrule

 
 \parbox[t]{5mm}{\multirow{30}{*}{\rotatebox[origin=c]{90}{Greedy Search}}} & \multirow{6}{*}{ALM(u)} & Bloom3B & Il a construit un interphone WiFi, il a dit. & Il a construit un interphone WiFi, il a dit. \\
 &  & Bloom7B & Il a construit un interphone WiFi, il a dit. & Il a construit un interphone sans fil, il a dit. \\
 &  & Llama2-7B & Il a construit un interphone WiFi, il a dit. & Il a construit un interphone WiFi, il a dit. \\
 &  & Llama2-7BChat & Il a construit un timbre WiFi. & Il a construit un interphone Wi-Fi, il a dit. \\
 &  & XGLM2.9B & Il a construit un WiFi, il a dit. & Il a construit un WiFi, il a dit. \\
 &  & XGLM7.5B & Il a construit un interphone WiFi, il a dit. & He a wifi porte-clefs, il a dit. \\
  \cline{2-5}
 & \multirow{6}{*}{ALM(x)} & Bloom3B & Il a construit un interphone WiFi, il a dit. & Il a construit un interphone WiFi, il a dit. \\
 &  & Bloom7B & Il a construit un interphone WiFi, il a dit. & Il a construit un interphone sans fil, il a dit. \\
 &  & Llama2-7B & Il a construit un interphone WiFi, il a dit. & Il a fait un interphone WiFi, il a dit. \\
 &  & Llama2-7BChat & Il a construit un timbre WiFi. & Il a construit un interphone Wi-Fi, il a dit. \\
 &  & XGLM2.9B & \textless{}EMPTY GENERATION\textgreater{} & Il a construit un WiFi, il a dit. \\
 &  & XGLM7.5B & \textless{}EMPTY GENERATION\textgreater{} & Il a construit un WiFi porte-clefs, il a dit. \\
 \cline{2-5}
 & \multirow{6}{*}{PMI(u)} & Bloom3B & Il a construit un interphone WiFi, il a dit. & Il a construit un interphone WiFi, il a dit. \\
 &  & Bloom7B & Il a construit un interphone WiFi, il a dit. & Il a construit un interphone WiFi, il a dit. \\
 &  & Llama2-7B & Il a construit un interphone WiFi, il a dit. & Il a construit un interphone WiFi, il a dit. \\
 &  & Llama2-7BChat & Il a construit un timbre WiFi. & Il a construit un interphone Wi-Fi, il a dit. \\
 &  & XGLM2.9B & Il a construit un WiFi, il a dit. & Il a construit un WiFi porte-fenêtre, il a dit. \\
 &  & XGLM7.5B & Il a construit un interphone WiFi, il a dit. & Il a construit un WiFi porte-clefs, il a dit. \\
  \cline{2-5}
 & \multirow{6}{*}{PMI(x)} & Bloom3B & Il a construit un appareil de sonnerie WiFi, il a dit. & Il a construit un interphone sans fil, il a dit. \\
 &  & Bloom7B & Il a construit un interphone WiFi, il a dit. & Il a construit un interphone sans fil, il a dit. \\
 &  & Llama2-7B & Il a construit un appareil WiFi pour son interphone, il a dit. & Il a fait un interphone WiFi, il a dit. \\
 &  & Llama2-7BChat & Il a construit un timbre WiFi. & Il a construit un interphone Wi-Fi, il a dit. \\
 &  & XGLM2.9B & Il a construit un WiFi, il a dit. & Il a construit un WiFi porte-fenêtre, il a dit. \\
 &  & XGLM7.5B & Il a construit un WiFi porte-clefs, il a dit. & Il a construit un WiFi porte-clefs, il a dit. \\
  \cline{2-5}
 & \multirow{6}{*}{base} & Bloom3B & Il a construit un interphone WiFi, il a dit. & Il a construit un interphone WiFi, il a dit. \\
 &  & Bloom7B & Il a construit un interphone WiFi, il a dit. & Il a construit un interphone sans fil, il a dit. \\
 &  & Llama2-7B & Il a construit un interphone WiFi, il a dit. & Il a fait un interphone WiFi, il a dit. \\
 &  & Llama2-7BChat & Il a construit un timbre WiFi. & Il a construit un interphone Wi-Fi, il a dit. \\
 &  & XGLM2.9B & Il a construit un WiFi, il a dit. & Il a construit un WiFi, il a dit. \\
 &  & XGLM7.5B & \textless{}EMPTY GENERATION\textgreater{} & \textless{}EMPTY GENERATION\textgreater{} \\
 \bottomrule
\end{tabular}}
\label{tab:trans_ex_masterful}
\caption{Example translation of a French sentence from various generation settings and models with the different prompts.}
\end{table*}

\newcommand{\emptygen}{\textless{}EMPTY GENERATION\textgreater{}}

\begin{table*}[]
    \centering
    \begin{tabular}{lll}
    \toprule
    Objective & Model Translation & Missing Entities \\
    \toprule
Source & \resizecol{0.6}{Dr. Ehud Ur, professor of medicine at Dalhousie University in Halifax, Nova Scotia and chair of the clinical and scientific division of the Canadian Diabetes Association cautioned that the research is still in its early days.} & \\
Target & \resizecol{0.6}{Le Dr Ehud Ur, professeur de médecine à l'Université Dalhousie de Halifax (Nouvelle-Écosse) et président de la division clinique et scientifique de l'Association canadienne du diabète, a averti que la recherche en était encore à ses débuts.} & \\
\toprule
Base & \emptygen & Ehud Ur, Halifax \\
\midrule
PMI($x$) & \resizecol{0.6}{Dr. Ehud Ur, professeur de médecine à l'Université Dalhousie de Halifax, au Canada et président de la division clinique et scientifique de la Société canadienne du diabète, a souligné que les résultats de cette étude sont encore très précoces.} & \\
\midrule
ALM($x$) & \resizecol{0.6}{Dr. Ehud Ur, professeur de médecine à l'Université Dalhousie de Halifax, en Nouvelle-Écosse et président de la division de la recherche clinique et scientifique de la Société canadienne du diabète, a déclaré que le travail est encore en cours.} & \\
\midrule
PMI($u$) & \resizecol{0.6}{Dr. Ehud Ur, professeur de médecine à l'Université Dalhousie de Halifax, en Nouvelle-Écosse et président de la division clinique et scientifique de la Société canadienne du diabète, avertit que le travail est encore dans ses premières étapes.} & \\
\midrule
ALM($u$) & \resizecol{0.6}{Dr. Ehud Ur, professeur de médecine à l'Université Dalhousie de Halifax, en Nouvelle-Écosse et président de la division de la recherche clinique et scientifique de la Société canadienne du diabète, a déclaré que le travail est encore en cours.} & \\
\bottomrule
    \end{tabular}
    \caption{Example translations from English to French across decoding methods for XGLM 7.5B model with greedy search. The empty generation from the default decoding objective is considered a failure to translate.}
    \label{tab:failure_ex}
\end{table*}

\begin{table*}
\centering

\resizebox{\textwidth}{!}{
\begin{tabular}{llllllllllll}
\toprule
 &  & \multicolumn{5}{c}{\textbf{Greedy}} & \multicolumn{5}{c}{\textbf{Beam Search (B=5)}} \\
   \cmidrule(lr){3-7} \cmidrule(lr){8-12} 

  &  & base & $\text{PMI}_u$ & $\text{PMI}_x$ & $\text{ALM}_u$ & $\text{ALM}_x$ & base & $\text{PMI}_u$ & $\text{PMI}_x$ & $\text{ALM}_u$ & $\text{ALM}_x$  \\

\midrule
\multirow{6}{*}{en $\rightarrow$ fr} & XGLM2.9B & 66.59 & \textbf{71.86} & 69.98 & {\ul 71.73} & 71.32 & 64.75 & 59.99 & 62.08 & 75.07 & \textbf{75.96} \\
 & XGLM7.5B & 68.26 & 74.82 & 73.54 & 74.77 & \textbf{75.42} & 56.49 & 66.31 & 63.27 & \textbf{75.85} & 75.47 \\
 & Bloom 3B & 77.53 & \textbf{79.67} & {\ul 79.60} & 78.72 & 79.11 & {\ul 80.23} & 78.59 & 78.60 & 79.91 & \textbf{80.39} \\
 & Bloom 7B & 80.98 & {\ul 81.70} & \textbf{81.79} & {\ul 81.68} & 81.03 & 81.76 & 81.70 & 81.19 & \textbf{82.05} & {\ul 81.95} \\
 & Llama 7B & {\ul 83.86} & {\ul 83.88} & \textbf{83.93} & {\ul 83.77} & 83.49 & 84.68 & 82.99 & 82.87 & {\ul 84.89} & \textbf{85.00} \\
 & Llama-chat 7B & 82.62 & 82.51 & 82.60 & \textbf{82.88} & {\ul 82.79} & 83.52 & 80.51 & 81.52 & 83.46 & \textbf{83.72} \\
 \midrule
\multirow{6}{*}{en $\rightarrow$ pt} & XGLM2.9B & 54.05 & 64.26 & 63.21 & 71.66 & \textbf{73.26} & 49.71 & 50.76 & 66.00 & 73.82 & \textbf{77.22} \\
 & XGLM7.5B & 61.15 & 75.27 & 71.81 & \textbf{78.78} & 77.98 & 45.69 & 60.71 & 56.25 & \textbf{78.16} & 77.78 \\
 & Bloom 3B & 82.76 & \textbf{83.92} & 83.52 & 83.34 & 83.54 & {\ul 84.25} & 83.33 & 83.86 & 84.02 & \textbf{84.30} \\
 & Bloom 7B & 83.85 & \textbf{84.89} & 84.45 & {\ul 84.76} & 84.42 & {\ul 85.28} & 85.17 & 84.07 & \textbf{85.41} & 84.53 \\
 & Llama 7B & {\ul 85.85} & {\ul 85.91} & \textbf{86.04} & {\ul 85.89} & {\ul 85.86} & \textbf{86.62} & 85.45 & 85.15 & 86.36 & {\ul 86.60} \\
 & Llama-chat 7B & 83.49 & 83.90 & 83.85 & {\ul 84.21} & \textbf{84.33} & 84.35 & 82.21 & 82.40 & {\ul 84.96} & \textbf{85.02} \\
 \midrule
\multirow{6}{*}{en $\rightarrow$ de} & XGLM2.9B & 64.91 & 68.40 & 67.46 & 67.67 & \textbf{68.71} & 62.01 & 57.58 & 60.97 & 71.60 & \textbf{73.17} \\
 & XGLM7.5B & 59.22 & 70.16 & 67.44 & 73.60 & \textbf{74.00} & 45.55 & 58.15 & 53.88 & {\ul 73.59} & \textbf{73.63} \\
 & Bloom 3B & 53.38 & \textbf{53.92} & 53.48 & 53.45 & 46.99 & 56.39 & 55.62 & 51.55 & \textbf{57.18} & 49.09 \\
 & Bloom 7B & 52.74 & \textbf{54.49} & 53.86 & 54.00 & 52.38 & 56.33 & 56.75 & 40.87 & \textbf{57.63} & 54.03 \\
 & Llama 7B & 81.45 & 81.30 & 81.44 & 81.50 & \textbf{81.98} & 81.70 & 80.10 & 80.43 & 81.81 & \textbf{83.17} \\
 & Llama-chat 7B & 77.94 & 77.99 & 78.07 & 78.64 & \textbf{79.02} & 79.59 & 75.76 & 76.91 & {\ul 80.19} & \textbf{80.31} \\
 \bottomrule
\end{tabular}}
\caption{Translation performance on FLORES with greedy decoding and beam search ($B=5$). \textbf{Scores are reported with COMET-22} \citep{rei-etal-2022-comet}, where higher is better. ``base" refers to default maximum likelihood decoding. The best scores are bolded.}
\label{tab:results-comet}
\end{table*}

\begin{table*}
\centering
\resizebox{\textwidth}{!}{
\begin{tabular}{llllllllllll}
\toprule
 &  & \multicolumn{5}{c}{\textbf{Greedy}} & \multicolumn{5}{c}{\textbf{Beam Search (B=5)}} \\
   \cmidrule(lr){3-7} \cmidrule(lr){8-12} 

  &  & base & $\text{PMI}_u$ & $\text{PMI}_x$ & $\text{ALM}_u$ & $\text{ALM}_x$ & base & $\text{PMI}_u$ & $\text{PMI}_x$ & $\text{ALM}_u$ & $\text{ALM}_x$  \\

\midrule
\multirow{6}{*}{en $\rightarrow$ fr} & XGLM2.9B & 65.18 & \textbf{70.03} & 68.55 & 69.59 & {\ul 69.97} & 63.27 & \textit{61.07} & 58.15 & 74.24 & \textbf{75.49} \\
 & XGLM7.5B & \textit{68.62} & \textit{76.30} & \textit{75.43} & \textit{76.46} & \textit{\textbf{77.25}} & 56.28 & \textit{68.47} & 62.82 & \textit{77.41} & \textit{\textbf{79.03}} \\
 & Bloom 3B & \textit{79.80} & \textit{\textbf{80.81}} & \textit{80.29} & \textit{79.88} & {\ul \textit{80.65}} & \textit{81.81} & \textit{80.67} & \textit{80.49} & \textit{80.97} & \textit{\textbf{82.58}} \\
 & Bloom 7B & \textit{82.77} & \textit{\textbf{83.66}} & \textit{83.45} & \textit{82.97} & \textit{82.76} & \textit{\textbf{84.62}} & \textit{83.76} & \textit{83.33} & \textit{84.08} & \textit{84.23} \\
 & Llama 7B & \textbf{83.75} & 83.52 & {\ul 83.58} & 83.24 & 83.41 & \textit{84.98} & 82.69 & 82.24 & 83.91 & \textit{\textbf{85.37}} \\
 & Llama-chat 7B & {\ul 82.41} & 82.24 & 82.37 & {\ul 82.44} & \textbf{82.61} & {\ul \textit{83.79}} & \textit{80.51} & \textit{81.56} & \textit{83.55} & \textit{\textbf{83.89}} \\
 \midrule
\multirow{6}{*}{en $\rightarrow$ pt} & XGLM2.9B & 52.49 & \textit{68.93} & 62.44 & 70.79 & \textbf{72.53} & \textit{49.86} & \textit{53.15} & 48.72 & \textit{75.37} & \textit{\textbf{77.77}} \\
 & XGLM7.5B & 59.65 & \textit{76.81} & 71.68 & \textit{\textbf{80.63}} & {\ul \textit{80.54}} & 45.34 & \textit{63.58} & 52.88 & \textit{80.62} & \textit{\textbf{81.19}} \\
 & Bloom 3B & \textit{83.48} & \textit{\textbf{83.98}} & {\ul \textit{83.82}} & \textit{83.69} & {\ul \textit{83.94}} & {\ul \textit{85.13}} & \textit{84.30} & \textit{84.20} & \textit{84.88} & \textit{\textbf{85.21}} \\
 & Bloom 7B & \textit{85.58} & {\ul \textit{85.79}} & \textit{\textbf{85.84}} & \textit{85.35} & \textit{85.36} & \textit{\textbf{86.74}} & \textit{86.07} & \textit{86.00} & {\ul \textit{86.71}} & \textit{86.51} \\
 & Llama 7B & {\ul 85.83} & 85.51 & {\ul 85.87} & 85.64 & \textit{\textbf{85.96}} & \textit{86.71} & 85.16 & 84.81 & 86.31 & \textit{\textbf{87.19}} \\
 & Llama-chat 7B & {\ul \textit{84.55}} & {\ul \textit{84.51}} & {\ul \textit{84.44}} & \textit{84.41} & \textit{\textbf{84.62}} & \textit{85.30} & \textit{83.27} & \textit{82.93} & \textit{85.46} & \textit{\textbf{85.66}} \\
 \midrule
\multirow{6}{*}{en $\rightarrow$ de} & XGLM2.9B & \textit{65.13} & 68.00 & \textit{67.50} & 66.78 & \textit{\textbf{69.19}} & \textit{63.18} & \textit{58.86} & 58.70 & 71.39 & \textit{\textbf{73.83}} \\
 & XGLM7.5B & 59.08 & \textit{70.94} & \textit{68.29} & \textit{73.84} & \textit{\textbf{74.76}} & 45.17 & \textit{58.94} & \textit{54.96} & 73.16 & \textit{\textbf{75.83}} \\
 & Bloom 3B & 47.26 & 48.06 & 46.70 & \textbf{48.52} & 44.61 & 52.73 & 51.06 & 47.44 & \textbf{53.33} & 48.72 \\
 & Bloom 7B & 51.27 & \textbf{54.10} & 52.46 & 53.77 & 51.94 & 55.96 & 54.38 & \textit{52.15} & \textit{\textbf{57.80}} & \textit{56.07} \\
 & Llama 7B & 81.02 & 80.68 & 81.31 & 80.50 & \textbf{81.64} & \textit{82.02} & \textit{80.26} & 80.31 & 80.83 & \textit{\textbf{83.45}} \\
 & Llama-chat 7B & \textit{\textbf{78.40}} & 77.99 & {\ul \textit{78.30}} & {\ul 78.37} & {\ul 78.35} & \textit{80.35} & \textit{76.50} & 76.87 & \textit{80.28} & \textit{\textbf{80.70}} \\
 \bottomrule
\end{tabular}}
\caption{Translation performance on FLORES with greedy decoding and beam search ($B=5$) \textbf{with the ``masterful'' prompt}. \textbf{Scores are reported with COMET-22} \citep{rei-etal-2022-comet}, where higher is better. ``base" refers to default maximum likelihood decoding. The best scores are bolded. Scores that are better than when using the basic prompt are italicized.}
\label{tab:results-comet}
\end{table*}

\begin{table*}
\resizebox{\textwidth}{!}{
\begin{tabular}{llllllllllll}
\toprule
 &  & \multicolumn{5}{c}{\textbf{Greedy}} & \multicolumn{5}{c}{\textbf{Beam Search (B=5)}} \\
   \cmidrule(lr){3-7} \cmidrule(lr){8-12} 

  &  & base & $\text{PMI}_u$ & $\text{PMI}_x$ & $\text{ALM}_u$ & $\text{ALM}_x$ & base & $\text{PMI}_u$ & $\text{PMI}_x$ & $\text{ALM}_u$ & $\text{ALM}_x$  \\

\midrule
\multirow{6}{*}{en $\rightarrow$ fr} & XGLM2.9B & 31.20 & \textbf{18.26} & 25.52 & 23.35 & 59.11 & 38.41 & 40.88 & 48.12 & \textbf{14.38} & 16.92 \\
 & XGLM7.5B & 29.72 & \textbf{14.53} & 20.28 & 18.45 & 55.43 & 53.22 & 27.41 & 37.70 & 18.49 & \textbf{17.11} \\
 & Bloom 3B & 20.58 & \textbf{13.65} & 15.91 & 15.86 & 21.90 & 13.55 & \textbf{11.27} & 16.76 & 11.79 & 16.29 \\
 & Bloom 7B & 13.05 & \textbf{9.12} & 10.69 & 11.58 & 17.71 & 9.42 & \textbf{7.60} & 11.38 & 9.35 & 13.04 \\
 & Llama 7B & 9.71 & \textbf{7.34} & 9.57 & 9.28 & 9.59 & 6.93 & \textbf{5.98} & 9.45 & 7.48 & 7.15 \\
 & Llama-chat 7B & 10.11 & \textbf{9.35} & 9.77 & 9.54 & 9.96 & 9.44 & \textbf{7.72} & 11.62 & 8.78 & 9.21 \\
 \midrule
\multirow{6}{*}{en $\rightarrow$ pt} & XGLM2.9B & 52.01 & 32.37 & 37.22 & \textbf{19.95} & 20.10 & 63.81 & 55.19 & 41.95 & 17.18 & \textbf{15.56} \\
 & XGLM7.5B & 47.71 & 18.47 & 28.26 & \textbf{12.39} & 18.67 & 72.63 & 46.02 & 56.46 & \textbf{13.62} & 17.05 \\
 & Bloom 3B & 12.41 & 9.62 & 11.75 & \textbf{9.59} & 14.51 & \textbf{9.08} & 9.94 & 13.53 & 9.31 & 12.86 \\
 & Bloom 7B & 10.17 & \textbf{8.94} & 9.60 & 9.65 & 11.15 & 6.79 & \textbf{5.62} & 10.12 & 8.80 & 9.42 \\
 & Llama 7B & \textbf{3.60} & 4.48 & 3.76 & 3.70 & 4.63 & 3.60 & \textbf{3.03} & 5.40 & 4.01 & 6.07 \\
 & Llama-chat 7B & 6.22 & \textbf{5.30} & 5.68 & \textbf{5.30} & \textbf{5.30} & 6.17 & 6.10 & 7.41 & \textbf{5.30} & \textbf{5.30} \\
 \midrule
\multirow{6}{*}{en $\rightarrow$ de}  & XGLM2.9B & 26.49 & \textbf{16.53} & 19.32 & 19.97 & 21.55 & 37.05 & 38.75 & 47.19 & \textbf{14.20} & 17.41 \\
 & XGLM7.5B & 43.85 & 21.29 & 27.13 & \textbf{13.59} & 15.18 & 67.14 & 36.51 & 53.01 & \textbf{16.21} & 16.58 \\
 & Bloom 3B & 12.01 & \textbf{8.67} & 10.55 & 11.41 & 22.23 & 7.66 & 7.94 & 19.19 & \textbf{7.06} & 19.17 \\
 & Bloom 7B & 19.64 & \textbf{12.92} & 16.18 & 16.26 & 20.99 & 18.81 & \textbf{11.39} & 20.87 & 13.82 & 18.90 \\
 & Llama 7B & \textbf{5.05} & 5.41 & 6.30 & 5.11 & 7.63 & 5.65 & 5.51 & 8.04 & \textbf{4.64} & 6.51 \\
 & Llama-chat 7B & 9.12 & 8.47 & 8.98 & 7.08 & \textbf{6.88} & 7.83 & 7.97 & 9.52 & \textbf{6.48} & 7.01 \\
\bottomrule
\end{tabular}}
\caption{Translation performance on FLORES with greedy decoding and beam search ($B=5$). Scores are reported for the \textbf{missing entity rate (MER)}. A lower score is better. ``base" refers to default maximum likelihood decoding. The best scores are bolded.}
\label{tab:results-MER}
\end{table*}

\begin{table*}
\resizebox{\textwidth}{!}{
\begin{tabular}{llllllllllll}
\toprule
 &  & \multicolumn{5}{c}{\textbf{Greedy}} & \multicolumn{5}{c}{\textbf{Beam Search (B=5)}} \\
   \cmidrule(lr){3-7} \cmidrule(lr){8-12} 

  &  & base & $\text{PMI}_u$ & $\text{PMI}_x$ & $\text{ALM}_u$ & $\text{ALM}_x$ & base & $\text{PMI}_u$ & $\text{PMI}_x$ & $\text{ALM}_u$ & $\text{ALM}_x$  \\

\midrule
\multirow{6}{*}{en $\rightarrow$ fr} & XGLM2.9B & 11.07 & 0.89 & 3.95 & \textbf{0.00} & \multicolumn{1}{c|}{46.08} & 19.86 & 6.13 & 9.78 & \textbf{0.00} & \textbf{0.00} \\
 & XGLM7.5B & 14.33 & 0.79 & 3.66 & \textbf{0.00} & \multicolumn{1}{c|}{45.63} & 36.86 & 13.83 & 21.34 & \textbf{0.10} & 0.20 \\
 & Bloom 3B & 3.26 & 0.10 & 0.79 & \textbf{0.00} & \multicolumn{1}{c|}{2.45} & 0.99 & 0.30 & 3.56 & \textbf{0.00} & 1.09 \\
 & Bloom 7B & 1.48 & 0.10 & 0.40 & \textbf{0.00} & \multicolumn{1}{c|}{2.14} & 0.40 & 0.10 & 1.09 & \textbf{0.00} & 0.89 \\
 & Llama 7B & 0.10 & \textbf{0.00} & \textbf{0.00} & \textbf{0.00} & \multicolumn{1}{c|}{\textbf{0.00}} & 0.10 & \textbf{0.00} & 0.30 & \textbf{0.00} & 0.10 \\
 & Llama-chat 7B & 0.40 & \textbf{0.00} & 0.10 & \textbf{0.00} & \multicolumn{1}{c|}{\textbf{0.00}} & 0.49 & \textbf{0.00} & 0.30 & \textbf{0.00} & \textbf{0.00} \\ \midrule
\multirow{6}{*}{en $\rightarrow$ pt} & XGLM2.9B & 39.53 & 20.16 & 19.66 & 0.20 & \multicolumn{1}{c|}{\textbf{0.00}} & 50.30 & 26.38 & 7.41 & 1.28 & \textbf{0.69} \\
 & XGLM7.5B & 33.20 & 7.02 & 13.44 & \textbf{0.00} & \multicolumn{1}{c|}{\textbf{0.00}} & 63.44 & 30.63 & 41.11 & \textbf{0.69} & 0.99 \\
 & Bloom 3B & 1.58 & 0.20 & 0.89 & \textbf{0.00} & \multicolumn{1}{c|}{1.09} & 0.69 & 0.59 & 0.89 & \textbf{0.10} & 0.99 \\
 & Bloom 7B & 0.79 & \textbf{0.00} & 0.49 & \textbf{0.00} & \multicolumn{1}{c|}{0.89} & 0.49 & 0.20 & 0.69 & \textbf{0.00} & 0.79 \\
 & Llama 7B & \textbf{0.00} & \textbf{0.00} & \textbf{0.00} & \textbf{0.00} & \multicolumn{1}{c|}{\textbf{0.00}} & \textbf{0.00} & \textbf{0.00} & 0.10 & \textbf{0.00} & 0.10 \\
 & Llama-chat 7B & 1.38 & 0.20 & 0.69 & \textbf{0.00} & \multicolumn{1}{c|}{\textbf{0.00}} & 1.28 & 0.20 & 0.69 & \textbf{0.00} & \textbf{0.00} \\ \midrule
\multirow{6}{*}{en $\rightarrow$ de} & XGLM2.9B & 6.82 & 0.49 & 2.47 & \textbf{0.00} & \multicolumn{1}{c|}{\textbf{0.00}} & 17.00 & 5.34 & 7.81 & 0.10 & \textbf{0.00} \\
 & XGLM7.5B & 30.34 & 7.11 & 14.33 & \textbf{0.00} & \multicolumn{1}{c|}{\textbf{0.00}} & 58.40 & 25.59 & 39.53 & 0.49 & \textbf{0.20} \\
 & Bloom 3B & 1.98 & 0.20 & 0.30 & \textbf{0.00} & \multicolumn{1}{c|}{1.28} & 2.77 & 1.28 & 5.14 & \textbf{0.00} & 2.57 \\
 & Bloom 7B & 4.15 & 0.20 & 1.58 & \textbf{0.00} & \multicolumn{1}{c|}{2.67} & 3.95 & 0.89 & 1.68 & \textbf{0.00} & 3.75 \\
 & Llama 7B & \textbf{0.00} & \textbf{0.00} & \textbf{0.00} & \textbf{0.00} & \multicolumn{1}{c|}{\textbf{0.00}} & 0.20 & \textbf{0.00} & 0.10 & \textbf{0.00} & 0.30 \\
 & Llama-chat 7B & 1.88 & 0.59 & 1.28 & 0.10 & \multicolumn{1}{c|}{\textbf{0.00}} & 1.48 & 0.30 & 0.89 & 0.10 & \textbf{0.00} \\
\bottomrule
\end{tabular}}
\caption{Translation performance on FLORES with greedy decoding and beam search ($B=5$). Scores are reported for the \textbf{rate of empty generation (REG)}, or how often the model does not produce an output. A lower score is better. ``base" refers to default maximum likelihood decoding. The best scores are bolded.}
\label{tab:results-REG}
\end{table*}

\begin{table*}
\resizebox{\textwidth}{!}{
\begin{tabular}{llllllllllll}
\toprule
 &  & \multicolumn{5}{c}{\textbf{Greedy}} & \multicolumn{5}{c}{\textbf{Beam Search (B=5)}} \\
   \cmidrule(lr){3-7} \cmidrule(lr){8-12} 

  &  & base & $\text{PMI}_u$ & $\text{PMI}_x$ & $\text{ALM}_u$ & $\text{ALM}_x$ & base & $\text{PMI}_u$ & $\text{PMI}_x$ & $\text{ALM}_u$ & $\text{ALM}_x$  \\

\midrule
\multirow{6}{*}{en $\rightarrow$ fr} & XGLM2.9B & 33.49 & \textbf{20.78} & 28.62 & \textit{23.95} & \textit{26.91} & \textit{40.45} & \textit{39.43} & \textit{51.68} & \textit{\textbf{13.02}} & 17.67 \\
 & XGLM7.5B & \textit{\textbf{27.16}} & \textit{\textbf{12.32}} & \textit{16.66} & \textit{12.70} & \textit{15.09} & \textit{48.74} & \textit{24.56} & \textit{38.71} & \textit{12.76} & \textit{\textbf{11.23}} \\
 & Bloom 3B & \textit{17.13} & \textit{\textbf{12.99}} & \textit{14.64} & \textit{15.47} & \textit{19.93} & \textit{12.78} & \textit{\textbf{9.19}} & \textit{14.26} & \textit{9.70} & \textit{15.49} \\
 & Bloom 7B & 13.94 & \textbf{11.22} & \textit{12.28} & \textit{11.49} & \textit{16.96} & 12.00 & \textbf{9.85} & 13.53 & 10.97 & 15.69 \\
 & Llama 7B & 10.16 & \textbf{9.31} & 10.16 & 9.95 & 12.23 & 7.17 & 6.24 & \textit{11.67} & \textit{\textbf{5.98}} & 9.35 \\
 & Llama-chat 7B & \textit{10.11} & \textit{\textbf{9.05}} & 9.92 & 10.30 & \textit{10.49} & \textit{9.40} & \textit{8.92} & \textit{10.99} & \textit{9.28} & \textit{\textbf{8.85}} \\
 \midrule
\multirow{6}{*}{en $\rightarrow$ pt} & XGLM2.9B & \textit{60.08} & \textit{25.07} & 43.90 & \textbf{20.35} & 20.73 & \textit{64.63} & \textit{53.92} & \textit{71.57} & \textit{15.41} & \textit{\textbf{13.11}} \\
 & XGLM7.5B & \textit{48.95} & \textit{16.13} & \textit{27.09} & \textit{10.12} & \textit{\textbf{9.81}} & \textit{73.10} & \textit{41.05} & \textit{59.08} & \textit{\textbf{11.82}} & \textit{13.76} \\
 & Bloom 3B & \textit{\textbf{10.49}} & \textit{\textbf{9.93}} & \textit{10.47} & \textit{9.98} & \textit{13.13} & \textit{9.16} & \textit{\textbf{8.02}} & \textit{12.68} & \textit{8.33} & \textit{10.58} \\
 & Bloom 7B & \textit{10.81} & \textit{\textbf{8.86}} & 10.35 & 11.15 & 13.20 & \textbf{7.21} & \textit{7.49} & \textit{9.45} & 9.24 & 11.04 \\
 & Llama 7B & 6.58 & \textbf{4.99} & 7.00 & 5.50 & 7.66 & 5.04 & \textbf{3.86} & 7.87 & \textit{4.73} & \textit{5.30} \\
 & Llama-chat 7B & 8.10 & 6.92 & 7.59 & \textbf{6.46} & \textit{6.87} & \textit{\textbf{5.68}} & 6.38 & 10.37 & 5.94 & 5.94 \\
 \midrule
 \multirow{6}{*}{en $\rightarrow$ de} & XGLM2.9B & 27.03 & \textbf{17.80} & \textit{21.04} & \textit{18.78} & \textit{22.91} & \textit{31.87} & \textit{37.26} & \textit{45.13} & \textit{\textbf{8.20}} & \textit{15.93} \\
 & XGLM7.5B & \textit{43.32} & \textit{19.15} & \textit{27.27} & \textit{\textbf{11.29}} & \textit{11.54} & 69.09 & \textit{39.85} & \textit{52.79} & \textit{\textbf{11.05}} & \textit{12.79} \\
 & Bloom 3B & 15.47 & \textbf{13.06} & 17.03 & 13.64 & 25.62 & 10.57 & \textit{12.32} & \textit{\textbf{17.07}} & \textit{\textbf{10.33}} & \textit{17.83} \\
 & Bloom 7B & 26.25 & \textbf{15.51} & \textit{23.45} & \textit{15.69} & \textit{24.11} & \textit{18.57} & 13.72 & \textit{23.18} & \textit{\textbf{10.60}} & 21.00 \\
 & Llama 7B & 5.74 & \textit{5.80} & \textit{\textbf{5.63}} & 6.42 & \textit{9.87} & \textit{5.12} & 6.07 & 9.50 & \textbf{4.96} & 7.86 \\
 & Llama-chat 7B & \textit{\textbf{8.00}} & 8.51 & 9.29 & 8.48 & 10.41 & \textit{7.92} & \textit{\textbf{7.39}} & 11.99 & 7.83 & 8.36 \\
\bottomrule
\end{tabular}}
\caption{Translation performance on FLORES with greedy decoding and beam search ($B=5$) \textbf{with the ``masterful'' prompt}. Scores are reported for the \textbf{missing entity rate (MER)}. A lower score is better. ``base" refers to default maximum likelihood decoding. The best scores are bolded. Scores that are better than when using the basic prompt are italicized.}
\label{tab:results-MER-masterful}
\end{table*}

\begin{table*}
\resizebox{\textwidth}{!}{
\begin{tabular}{llllllllllll}
\toprule
 &  & \multicolumn{5}{c}{\textbf{Greedy}} & \multicolumn{5}{c}{\textbf{Beam Search (B=5)}} \\
   \cmidrule(lr){3-7} \cmidrule(lr){8-12} 

  &  & base & $\text{PMI}_u$ & $\text{PMI}_x$ & $\text{ALM}_u$ & $\text{ALM}_x$ & base & $\text{PMI}_u$ & $\text{PMI}_x$ & $\text{ALM}_u$ & $\text{ALM}_x$  \\

\midrule
\multirow{6}{*}{en $\rightarrow$ fr} & XGLM2.9B & \textit{10.77} & \textit{0.79} & \textit{3.66} & \textit{\textbf{0.00}} & \textit{\textbf{0.00}} & 21.05 & 6.23 & 10.28 & \underline{ 0.10} & \textit{\textbf{0.00}} \\
 & XGLM7.5B & 16.11 & 1.38 & 3.85 & \textit{\textbf{0.00}} & \textit{\textbf{0.00}} & 41.01 & \textit{13.83} & 25.79 & \underline{ 0.20} & \textit{\textbf{0.00}} \\
 & Bloom 3B & \textit{0.40} & \textit{\textbf{0.00}} & \underline{ \textit{0.20}} & \textit{\textbf{0.00}} & \textit{0.49} & \underline{ \textit{0.20}} & \textit{\textbf{0.00}} & \textit{0.49} & \textit{\textbf{0.00}} & \textit{0.40} \\
 & Bloom 7B & \textit{0.79} & \textit{\textbf{0.00}} & \underline{ \textit{0.20}} & \textit{\textbf{0.00}} & \textit{0.89} & \textit{\textbf{0.00}} & \underline{ \textit{0.10}} & \textit{0.40} & \textit{\textbf{0.00}} & \textit{0.69} \\
 & Llama 7B & \textit{\textbf{0.00}} & \textit{\textbf{0.00}} & \textit{\textbf{0.00}} & \textit{\textbf{0.00}} & \textit{\textbf{0.00}} & \textit{\textbf{0.00}} & \textit{\textbf{0.00}} & \textit{\textbf{0.00}} & \textit{\textbf{0.00}} & \textit{\textbf{0.00}} \\
 & Llama-chat 7B & \textit{\textbf{0.00}} & \textit{\textbf{0.00}} & \textit{\textbf{0.00}} & \textit{\textbf{0.00}} & \textit{\textbf{0.00}} & \underline{ \textit{0.20}} & \textit{\textbf{0.00}} & \textit{\textbf{0.00}} & \textit{\textbf{0.00}} & \textit{\textbf{0.00}} \\
 \midrule
\multirow{6}{*}{en $\rightarrow$ pt} & XGLM2.9B & 43.48 & \textit{7.21} & 21.64 & \textit{\textbf{0.10}} & \textbf{0.10} & 50.59 & \textit{22.43} & 37.06 & \textit{1.19} & \textit{\textbf{0.20}} \\
 & XGLM7.5B & 40.42 & 7.61 & 17.39 & \textit{\textbf{0.00}} & \underline{ 0.20} & 66.60 & 31.32 & 51.28 & 0.79 & \textit{\textbf{0.49}} \\
 & Bloom 3B & \textit{0.40} & \textit{\textbf{0.00}} & \underline{ \textit{0.20}} & \textit{\textbf{0.00}} & \textit{0.40} & \textbf{0.00} & \textbf{0.00} & \textit{0.20} & \textit{\textbf{0.00}} & \textit{0.49} \\
 & Bloom 7B & \textit{\textbf{0.00}} & \textit{\textbf{0.00}} & \textit{\textbf{0.00}} & \textit{\textbf{0.00}} & \textit{0.79} & \underline{ \textit{0.10}} & \textit{\textbf{0.00}} & \underline{ \textit{0.10}} & \textit{\textbf{0.00}} & \underline{ \textit{0.20}} \\
 & Llama 7B & \textit{\textbf{0.00}} & \textit{\textbf{0.00}} & \textit{\textbf{0.00}} & \textit{\textbf{0.00}} & \textit{\textbf{0.00}} & \textit{\textbf{0.00}} & \textit{\textbf{0.00}} & \textit{\textbf{0.00}} & \textit{\textbf{0.00}} & \textit{\textbf{0.00}} \\
 & Llama-chat 7B & \textit{\textbf{0.00}} & \textit{\textbf{0.00}} & \textit{\textbf{0.00}} & \textit{\textbf{0.00}} & \textit{\textbf{0.00}} & \underline{ \textit{0.20}} & \textit{\textbf{0.00}} & \textit{\textbf{0.00}} & \textit{\textbf{0.00}} & \textit{\textbf{0.00}} \\
 \midrule
\multirow{6}{*}{en $\rightarrow$ de} & XGLM2.9B & \textit{6.72} & \textit{0.49} & 2.67 & \textit{\textbf{0.00}} & \textit{\textbf{0.00}} & \textit{14.82} & 5.83 & 8.40 & \textit{\textbf{0.10}} & \textbf{0.10} \\
 & XGLM7.5B & 31.72 & 8.10 & 14.72 & \textit{\textbf{0.00}} & \textit{\textbf{0.00}} & 59.49 & 28.66 & \textit{38.93} & 0.69 & \textit{\textbf{0.20}} \\
 & Bloom 3B & \textit{0.59} & \underline{ \textit{0.10}} & \underline{ \textit{0.20}} & \textit{\textbf{0.00}} & \textit{0.30} & \textit{1.19} & \textit{0.30} & \textit{1.48} & \textit{\textbf{0.00}} & \textit{0.59} \\
 & Bloom 7B & 9.68 & 1.19 & 4.84 & \textbf{0.10} & 6.32 & 5.53 & 1.28 & 5.53 & \textit{\textbf{0.00}} & 5.34 \\
 & Llama 7B & \textit{\textbf{0.00}} & \textit{\textbf{0.00}} & \textit{\textbf{0.00}} & \textit{\textbf{0.00}} & \textit{\textbf{0.00}} & \textit{\textbf{0.00}} & \textit{\textbf{0.00}} & \textit{\textbf{0.00}} & \textit{\textbf{0.00}} & \textit{\textbf{0.00}} \\
 & Llama-chat 7B & \textit{0.30} & \textit{\textbf{0.00}} & \textit{0.30} & \textit{\textbf{0.00}} & \textit{\textbf{0.00}} & \underline{ \textit{0.20}} & \textit{\textbf{0.00}} & \textit{0.30} & \textit{\textbf{0.00}} & \textit{\textbf{0.00}} \\
\bottomrule
\end{tabular}}
\caption{Translation performance on FLORES with greedy decoding and beam search ($B=5$)  \textbf{with the ``masterful'' prompt}. Scores are reported for the \textbf{rate of empty generation (REG)}, or how often the model does not produce an output. A lower score is better. ``base" refers to default maximum likelihood decoding. The best scores are bolded and scores within 0.2 of the best are underlined. Scores that are better than when using the basic prompt are italicized.}
\label{tab:results-REG-masterful}
\end{table*}

\begin{table*}[]
\centering
\resizebox{\textwidth}{!}{
\begin{tabular}{llllllllllll}
\toprule
 &  & \multicolumn{5}{c}{\textbf{Greedy}} & \multicolumn{5}{c}{\textbf{Beam Search (B=5)}} \\
   \cmidrule(lr){3-7} \cmidrule(lr){8-12} 

  &  & base & $\text{PMI}_u$ & $\text{PMI}_x$ & $\text{ALM}_u$ & $\text{ALM}_x$ & base & $\text{PMI}_u$ & $\text{PMI}_x$ & $\text{ALM}_u$ & $\text{ALM}_x$  \\

\midrule
\multirow{6}{*}{en $\rightarrow$ fr}  & XGLM 2.9B & 17.4 & 19.9 & 18.8 & 19.0 & \textbf{20.4} & 15.5 &  \textit{13.8} & 10.9 & 22.7 & \textbf{24.8} \\
 & XGLM 7.5B &  \textit{22.5} &  \textit{27.2} &  \textit{26.6} &  \textit{27.4} & \textbf{28.1} &  \textit{13.9} &  \textit{22.3} &  \textit{17.9} &  \textit{28.7} & \textbf{30.8} \\
 & Bloom 3B &  \textit{30.3} & \textbf{31.5} &  \textit{31.3} &  \textit{30.1} &  \textit{31.0} &  \textit{33.5} &  \textit{31.6} &  \textit{31.5} &  \textit{32.9} & \textbf{35.3} \\
 & Bloom 7B &  \textit{35.3} &  \textit{35.7} & \textbf{36.4} &  \textit{35.6} &  \textit{35.3} & \textbf{39.1} &  \textit{36.5} &  \textit{36.7} &  \textit{38.6} &  \textit{38.6} \\
 & Llama 7B & \textbf{36.2} & 35.4 & 35.7 & 35.2 & 35.5 & \underline{ \textit{38.7}} & 34.7 & 32.4 & 36.1 & \textbf{38.7} \\
 & Llama-chat 7B & 33.7 & 33.5 & 33.7 & 33.7 & \textbf{34.1} &  \textit{35.4} & 32.5 &  \textit{32.8} &  \textit{35.1} & \textbf{35.7} \\
 \midrule
\multirow{6}{*}{en $\rightarrow$ pt}  & XGLM 2.9B & 7.1 &  \textit{15.7} & 11.9 & 15.8 & \textbf{19.0} & 5.1 &  \textit{7.5} & 4.8 & 18.6 & \textbf{23.9} \\
 & XGLM 7.5B & 13.6 &  \textit{25.5} &  \textit{22.3} & \textbf{28.1} &  \textit{27.9} & 3.5 &  \textit{16.3} & 8.1 &  \textit{29.7} & \textbf{30.8} \\
 & Bloom 3B &  \textit{30.3} & \underline{ \textit{30.6}} & \underline{ \textit{30.7}} & \underline{ \textit{30.6}} & \textbf{30.8} &  \textit{33.9} &  \textit{32.0} &  \textit{31.5} &  \textit{32.9} & \textbf{34.5} \\
 & Bloom 7B & \textbf{34.3} & \underline{ \textit{34.2}} & \underline{ \textit{34.3}} &  \textit{34.0} & \underline{ \textit{34.1}} & \underline{ \textit{37.4}} &  \textit{35.3} &  \textit{35.4} & \textbf{37.6} &  \textit{37.1} \\
 & Llama 7B & \underline{35.3} & 34.5 & \textbf{35.3} & 34.8 & \underline{35.1} &  {37.3} & 34.1 & 33.0 & 36.3 & \textbf{38.3} \\
 & Llama-chat 7B &  {33.8} &  \textit{33.5} &  \textit{33.7} &  \textit{33.7} & \textbf{34.0} &  \textit{34.9} &  \textit{32.6} &  \textit{32.3} &  \textit{34.9} & \textbf{35.2} \\
 \midrule
\multirow{6}{*}{en $\rightarrow$ de} & XGLM 2.9B & 11.8 & \underline{13.1} & 12.5 & 12.3 & \textbf{13.2} &  \textit{12.1} &  \textit{9.1} & 7.6 & 15.6 & \textbf{17.3} \\
 & XGLM 7.5B & 11.2 &  \textit{17.0} &  \textit{15.2} &  \textit{17.6} & \textbf{18.4} &  \textit{4.1} &  \textit{11.6} &  \textit{8.2} & 17.4 & \textbf{19.6} \\
 & Bloom 3B &  \textit{5.3} &  \textit{5.8} &  \textit{5.6} &  \textit{5.3} & \textbf{6.2} &  \textit{5.2} &  \textit{5.4} &  \textit{5.9} &  \textit{5.1} & \textbf{6.4} \\
 & Bloom 7B &  \textit{7.8} & \textbf{8.9} &  \textit{8.4} & \underline{ \textit{8.9}} &  \textit{8.3} &  \textit{9.3} &  \textit{8.8} &  \textit{8.0} & \textbf{9.6} & \underline{ \textit{9.6}} \\
 & Llama 7B & 24.7 & 24.3 & \textbf{25.0} & 23.9 & 24.6 &  \textit{25.7} & 23.8 & 22.9 & 24.4 & \textbf{27.3} \\
 & Llama-chat 7B & \textbf{22.2} & 21.9 & \underline{22.1} & \underline{22.1} & \underline{22.0} & 23.2 &  \textit{21.6} & 20.7 & 23.2 & \textbf{23.6} \\
\bottomrule
\end{tabular}}
\caption{Translation performance on FLORES with greedy decoding and beam search ($B=5$). \textbf{Scores are reported with SacreBLEU} \citep{post-2018-call}, where higher is better. ``base" refers to default maximum likelihood decoding. The best scores are bolded and scores within 0.2 of the best are underlined. The instructions used are  {"A <L1> phrase is provided. The masterful <L1> translator flawlessly translates the phrase into <L2>.''}, a verbose instruction phrase recommended by \citet{reynolds2021prompt}.}
\label{tab:results_masterful}
\end{table*}

\end{document}